\documentclass{bmvc2k}
\usepackage{booktabs}

\usepackage{graphicx}

\newcommand{\R}{\mathbb{R}} 

\usepackage{algorithm}
\usepackage{algpseudocode}
\usepackage{multirow}
\usepackage{amsfonts}
\usepackage{amsmath}
\usepackage{pifont}

\title{Revisiting LLM Adaptation for 3D CT Report Generation: A Study of Scaling and Diagnostic Priors}

\addauthor{Vanshali Sharma}{sharmavanshali@gmail.com}{1}
\addauthor{Andrea M. Bejar$^*$}{andrea.bejar@northwestern.edu}{1}
\addauthor{Halil Ertugrul Aktas$^*$}{halilertugrul.aktas@northwestern.edu}{1}
\addauthor{Quoc-Huy Trinh}{trnhquchuy@yahoo.com.vn}{3}
\addauthor{Debesh Jha}{debesh.jha@usd.edu}{2}
\addauthor{Gorkem Durak}{gorkem.durak@northwestern.edu}{1}
\addauthor{Ulas Bagci}{ulas.bagci@northwestern.edu}{1}

\addinstitution{
 Department of Radiology\\
 Northwestern University\\
 Chicago, USA
}
\addinstitution{
Department of Computer Science\\
University of South Dakota\\
 South Dakota, USA
}
\addinstitution{
 Department of Computer Science\\
 Aalto University, Finland
}

\runninghead{}{}

\begin{document}

\maketitle

\begin{abstract}
\def\thefootnote{*}\footnotetext{These authors contributed equally to this work.}\def\thefootnote{\arabic{footnote}}
Recent advances in multimodal learning, including large language models (LLMs) and vision-language models (VLMs), have demonstrated strong adaptability to natural images. However, extending their use to the medical domain, particularly for volumetric (3D) images, is challenging due to high computational complexity, volumetric dependencies and the semantic gap between visual features and clinical terminology. Naively fine-tuning LLMs on limited medical data often leads to overfitting and clinical hallucination, where linguistic fluency is prioritized over clinical factuality. In this study, we investigate parameter-efficient adaptation strategies for volumetric CT report generation and introduce \textbf{RAD3D-Prefix}, a lightweight diagnostic-prior conditioning framework that minimizes the need for extensive parameter training. This module integrates image embeddings with multi-label diagnostic classification logits, preserving critical clinical details while bridging the semantic gap. By keeping the LLM frozen, our method requires minimal trainable parameters and mitigates the risk of overfitting on small, domain-specific datasets. Through a systematic study spanning LLMs from 96.1M to 1.6B parameters, we find that fine-tuning is most beneficial for smaller LLMs, whereas freezing larger ($\approx$1B+) LLMs and training only lightweight projection layers provides a superior trade-off between performance, generalization, and computational efficiency. Across multiple automatic metrics and a clinical reader study, \textbf{RAD3D-Prefix} outperforms comparable parameter-efficient baselines and demonstrates strong out-of-domain generalization while using substantially fewer trainable parameters than fully fine-tuned alternatives. 
\footnote{The source code and models will be made public after the review process.} 

\end{abstract}

\section{Introduction}
\label{sec:intro}
Large Language Models (LLMs) are pre-trained on a massive amount of text, which allows them to generalize effectively and perform well on downstream tasks involving zero-shot learning~\cite{sanhmultitask, ramesh2021zero} and few-shot in-context learning~\cite{brown2020language,wei2022chain}. These remarkable properties of LLMs have inspired their adoption in various vision-based tasks for multimodal applications~\cite{guo2023images, li2024dtllm}. Most of these approaches concentrate on end-to-end training or fine-tuning using domain-specific image-text pairs ~\cite{li2022blip,kim2021vilt}. Specifically, multimodal models that work with medical images and text often rely on fine-tuning because the general data used to pre-train LLMs contains very few medical examples~\cite{e3d-gpt}.  Specialized biomedical LLMs have been developed such as \textbf{BioGPT} \cite{10.1093/bib/bbac409} and \textbf{BioMedLM} \cite{bolton2024biomedlm}, to address the limitations of general LLMs in the medical field. These models are pre-trained on extensive biomedical text corpora like \textbf{PubMed}~\cite{white2020pubmed} to improve their ability to recognize medical terminology. However, despite these advances, there is still a lack of a systematic approach to integrate LLMs in 3D medical imaging. Although the adoption of LLMs, including their frozen or fine-tuning paradigms has been extensively investigated in natural image-text settings \cite{llava, li2022blip}, their extension to volumetric 3D CT report generation, where critical diagnostic reasoning is crucial, has received comparatively limited attention.




\begin{figure}[t]
    \centering
    \includegraphics[width=0.7\columnwidth]{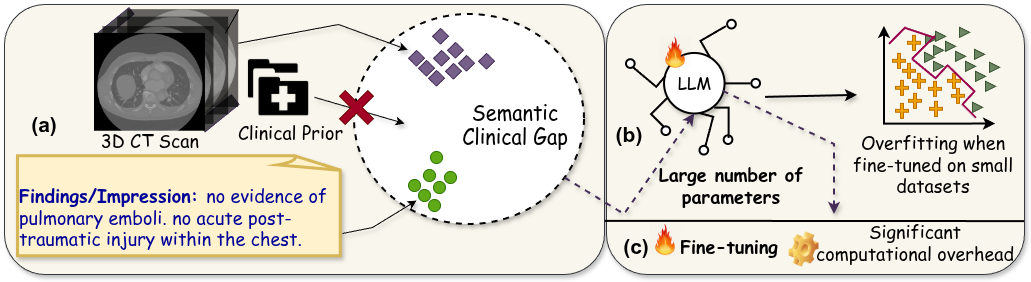}
    \caption{Three critical challenges in report generation: (a) Semantic Clinical Gap. (b) Clinical Hallucination. (c) Computational Inefficacy. }
    \label{fig:intro_3}
\end{figure}

\begin{figure*}[t]
    \centering
    \includegraphics[width=0.8\linewidth]{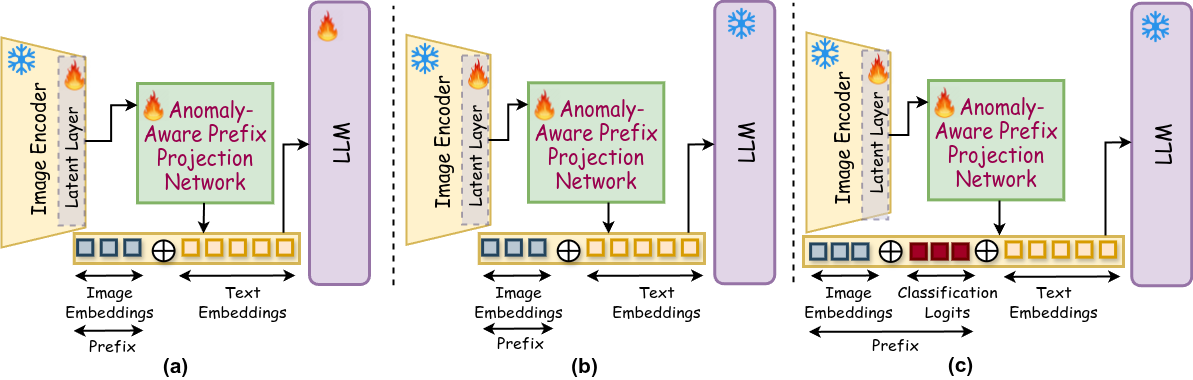}
    \caption{Three variations of the proposed projection module: (a) \textit{V-1}: prefix includes image embeddings and involves LLM's fine-tuning (suitable for smaller LLMs), (b) \textit{V-2}: prefix includes image embeddings with frozen LLM (suitable for larger LLMs), and (c) \textit{V-3}: prefix includes image embeddings, and classification logits with frozen LLM.}
    \label{fig:variants}
\end{figure*}

Generating clinical reports from 3D CT scans involves three critical issues (as illustrated in Fig~\ref{fig:intro_3}): 
(i) \textbf{\textit{Clinical Hallucination:}} Fine-tuning large models on limited medical image-text pairs often leads to overfitting and ``clinical hallucination'', where it prioritizes linguistic fluency over clinical factuality. Moreover, despite the growing adoption of LLMs in medical imaging, the trade-offs between freezing and fine-tuning models of different scales for 3D CT report generation remain poorly understood. (ii) \textbf{\textit{Computational Inefficiency:}} Fine-tuning process requires optimizing millions or billions of parameters, which creates significant computational overhead in a resource constrained clinical setting. (iii) \textbf{\textit{Semantic Clinical Gap:}} Unlike regular captions and X-ray reports, the 3D CT scan reports have long sequences and complex domain-specific language. Although 3D CT scans contain comprehensive diagnostic information compared to 2D images, absence of explicit clinical priors creates a semantic clinical gap between high-dimensional 3D visual features and medical terminologies. This issue persists even when using domain-specialized LLMs.

To address the above challenges, we conduct a systematic study of LLM adaptation strategies for 3D CT report generation across \textit{five} LLMs of varying parameter scales. Our study establishes a comprehensive  protocol for LLM training strategies and derives practical recommendations for choosing between freezing and fine-tuning paradigms for 3D CT report generation. Building on these findings, we propose \textit{\textbf{RAD3D-Prefix}}, a \textbf{\textit{lightweight anomaly-aware prefix conditioning framework}} that injects critical clinical priors into a frozen LLM through a prefix learning mechanism while allowing faster training with minimal trainable parameters. To this end, our method introduces a projection module that generates an \textit{anomaly-aware prefix}, which is a fixed-length embedding sequence that represents both image features and multi-label diagnostic classes. This lightweight approach allows LLMs to be used in a zero-shot learning setting. 



To systematically demonstrate the impact of our proposed approach, inspired by \cite{mokady2021clipcap, wang2023r2gengpt}, we investigated three different experiment setup (variant), as illustrated in Fig \ref{fig:variants}. In variant \textit{V-1} (Fig \ref{fig:variants} (a)), our projection network and LLMs are simultaneously trained with prefixes comprising of image embeddings alone, which are then concatenated with text embeddings. In variant \textit{V-2} (Fig \ref{fig:variants} (b)), the same prefix settings are used with frozen LLM. In variant \textit{V-3} (Fig \ref{fig:variants} (c)), we incorporate the diagnostic details by concatenating multi-abnormality classification logits with image prefix representation as projection input for frozen LLM. We examined these approaches using smaller models (with a few million parameters) and larger models ($\approx 1B$+ parameters). 
Unlike \cite{wang2023r2gengpt}, our work focuses on improving the core processing of 3D image embeddings and their projection as input to LLMs while preserving clinically significant multi-abnormality entity markers. Our main contributions are summarized below: 

\begin{itemize}
     
    \item  \textbf{First systematic study of frozen-vs-fine-tuned LLM scaling laws for 3D CT report generation:} We conducted extensive experiments on \textit{three} different model variations to determine the most effective approach for different prefix designs. Further, we performed comparison across \textit{five} LLMs from 96.1M to 1.6B parameters, with frozen vs fine-tuned setups. This provides actionable guidance (fine-tune $<$ 1B, freeze $\approx$1B+) that has not been studied in 3D medical imaging and contradicts prevailing natural image findings (LLaVA~\cite{llava} and BLIP-2~\cite{li2023blip}).

    \item \textbf{A lightweight anomaly-aware prefix projection module:} We propose a \textit{lightweight anomaly-aware prefix projection module} to generate clinical reports for 3D radiology images with minimal parameter training. In contrast to existing 2D image-based vision-language models (VLMs) \cite{medflamingo}, \textbf{RAD3D‑Prefix} aligns 3D image embeddings and anomaly logits with a frozen LLM. Thus, narrowing the semantic clinical gap, unlike natural image models \cite{10.5555/3666122.3668264, jin2024unified}, which face semantic gaps, especially when deployed in the medical domain. While the basic prefix learning concept exists~\cite{mokady2021clipcap}, we extend it to volumetric CT report generation by integrating visual features with diagnostic priors, yielding improved clinical relevance and parameter efficiency.
    
    \item \textbf{Ensure clinically relevant outputs using medical-specific metrics and a reader study by clinical experts: }We incorporated multi-anomaly classification logits to retain important clinical details in the generated reports. This explicitly exposes clinical concepts (e.g., effusion, consolidation) to the LLM. We further used medical-specific evaluation metrics to ensure diagnostic precision. Additionally, a reader study with two clinical experts shows that our model generates reports with higher clinical relevance than both the baseline and the variant without clinical priors.
    \item \textbf{Outperforms similar-sized and domain-specialized models, while performing comparable with larger models:} Our proposed method, despite minimal training, empirically outperforms existing techniques when using frozen LLMs with the same parameter count and specialized domain pre-training. The model also performed comparably to methods employing frozen LLMs with higher parameter count, supported by bootstrap analysis. Moreover, using the same vision encoder across all methods shows that the gains stem from the anomaly-aware prefix rather than a heavier backbone. 
\end{itemize}

\section{Related Work}
\subsection{Medical Report Generation}

Llava-Med \cite{llavamed}, Med-Flamingo \cite{medflamingo}, and Med-PaLM \cite{medpalm}  are the major models designed for medical report generation that use Vision-Language Models (VLMs) trained on extensive image-text datasets. However, these models have a key limitation: they can't process 3D medical images like CT and MRI scans because of the high complexity and computational costs involved. To address this, other models like,  CT2Rep \cite{ct2rep}, CT-AGRG \cite{ct-agrg}, E3D-GPT \cite{e3d-gpt}, and Med-2E3 \cite{med2e3} have been developed. These solutions capture global features from 3D images and use them as input for text decoders to generate reports. 

The \textbf{CT2Rep} model uses a 3D medical vision encoder to extract global features from CT images and integrate them into a language model for report generation, showing initial  effectivness. Building on this, the \textbf{CT-AGRG} model incorporates abnormality-guided recognition, which allows the framework to detect anomalies and generate corresponding medical report descriptions. In addition, \textbf{E3D-GPT} introduced a large-scale 3D medical image dataset and a 3D medical image foundation model based on MAE~\cite{he2022masked}, which enhances the representation of visual information in the overall vision-language model.


    While these methods have promising initial results, they still have significant limitations. They either fine-tune larger LLMs (E3D-GPT), use a simple linear projection of 3D radiology images (CT-AGRG), or fine-tune both the image encoder and text decoder (CT2Rep). This can lead to overfitting and a lack of proper alignment between visual and textual semantics. While recent work~\cite{chen2025large} has explored region-guided mechanisms, our work investigates parameter-efficient and diagnostic-prior conditioning for 3D CT report generation.

\subsection{Vision Projector in Large Vision Language Models (VLMs)}
Vision projectors are modules designed to project the context of visual data in the same way as text, which helps align the image and text spaces. Early methods like LLaVA~\cite{llava} proposed a simple feed-forward layer for this purpose, with promising initial results. Later, LLaMA 3.2~\cite{llama32} introduced a cross-attention mechanism to tackle this alignment challenge.

In the context of medical imaging, the LLaVA-Med model~\cite{llavamed} uses a simple MLP projection layer, similar to the original LLaVA model~\cite{llava} . This approach works well for 2D images but isn't effective for 3D radiology images. For report generation, models like Med-2E3~\cite{med2e3} and Red2RG~\cite{chen2024large} use adapters to help projectors encode volumetric data, but they still have high computational costs. This is because they use two encoders to generate visual information, which significantly increases overhead. Like our proposed method, the R2GenGPT~\cite{wang2023r2gengpt} model uses a frozen LLM with a visual alignment layer. However, its alignment layer is a simple linear projection, which can lead to feature alignment issues.



To address these challenges, we propose \textbf{RAD3D-Prefix}, which uses a transformer-based, anomaly-aware prefix module. This module leverages prefix projection to effectively manage the differences between the image and text embedding spaces. This approach improves computational efficiency and provides better alignment, all while keeping the LLM frozen.

\begin{figure*}[t]
    \centering
    \includegraphics[width=0.9\linewidth]{ 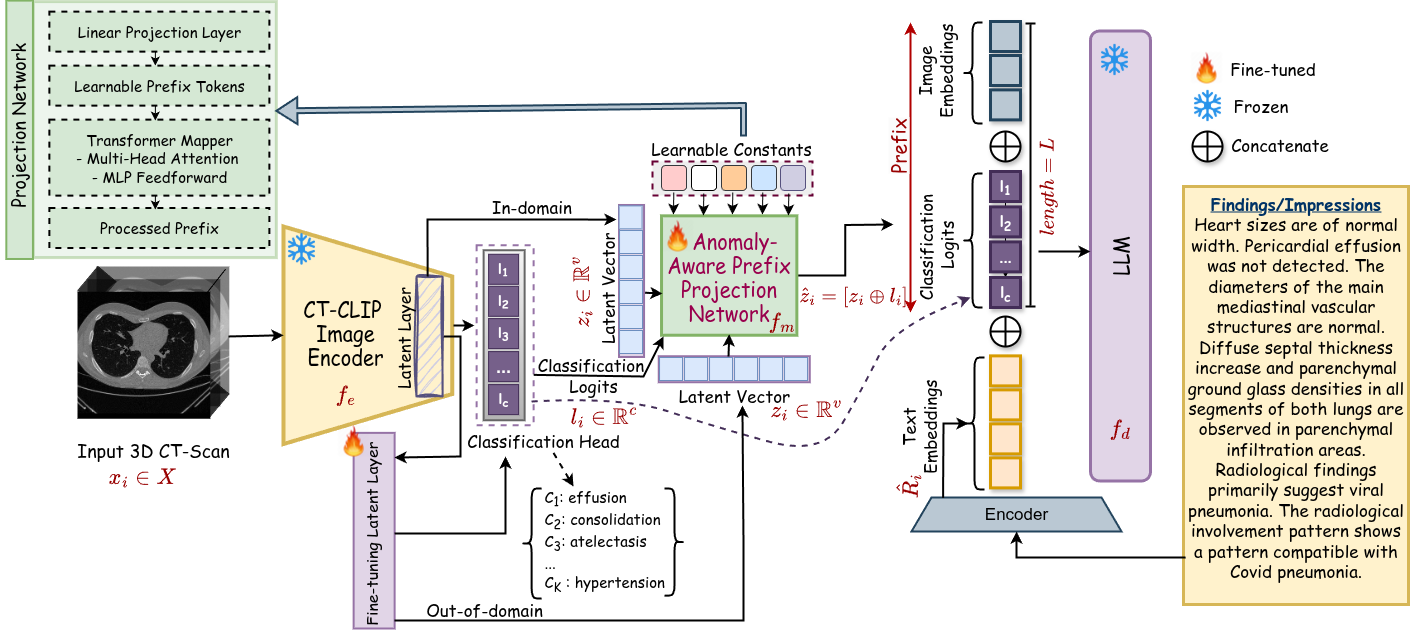}
    \caption{Overview of the proposed \textit{RAD3D-Prefix} model. The model aligns the image encoder's output and the classification logits to the text embedding space via a lightweight projection network.}
    \label{fig:overalframework}
\end{figure*}

\section{RAD3D-Prefix}
\label{sec:methodology}
In this section, we describe our proposed \textbf{RAD3D-Prefix} model for CT report generation. Given a 3D CT-scan image $x_i\in \mathbb{R}^{C\times D\times H\times W}$ where C, D, H and W denote the channel, depth, height, and width, respectively,  our framework aims to generate a patient-specific multi-sentence report impression/finding, $R$, in a clear and coherent manner. We leverage the strength of the vision foundation model and LLM. This approach allows for efficient report generation with only a few trainable parameters while still achieving enhanced outcomes and mitigating the previously mentioned limitations.   

\subsection{Overall Framework}
\textbf{RAD3D-Prefix} is presented in Fig. \ref{fig:overalframework}, comprising of a pretrained and frozen 3D image encoder $f_e$, a trainable transformer-based projection network $f_m$ and a frozen text decoder $f_d$. The encoder $f_e$ extracts the visual embedding that can be utilized by the decoder $f_d$ to generate a report $R_i=\{r_1,r_2,...,r_N\}\in \mathbb{V}$ where $\mathbb{V}$ is the vocabulary and $N$ is the length of the report. During this process, the lightweight network $f_m$ tries to project the visual embeddings obtained from $f_e$ into the $f_d$'s token space using a prefix learning mechanism, thus ensuring alignment between the two modalities.  For instance, given a dataset $D={(X,\mathcal{R})}$, we extract the visual embeddings $\mathbf{z}_i$ for image $x_i\in X$ using $f_e$:
\begin{equation}
    \mathbf{z}_i = f_e(x_i)\in \mathbb{R}^{v},
\end{equation}

To provide the model with crucial clinical context, our approach incorporates multi-anomaly classification logits, 
$\mathbf{l}_i\in \mathbb{R}^c$ (for $c$ anomaly labels). These logits are obtained from a separate, pre-trained classification head on the image encoder $f_e$. We directly fuse this high-level diagnostic information with the raw visual features by concatenating the image embeddings, $\mathbf{z}_i$, with the classification logits, $\mathbf{l}_i$. This concatenated vector, which combines both visual and clinical cues, is then used as input to our transformer-based projection network, $f_m$. The network transforms this combined input into a structured sequence of $L$-dimensional embedding vector, which act as a prefix for the LLM. Note that the \textbf{RAD3D-Prefix} is built on the \textit{V-3} variant configuration, where the prefix integrates both image embeddings and classification logits, and the LLM remains frozen. The other two proposed variants, \textit{V-1} and \textit{V-2} are presented in the subsequent sections for comparative analysis. 


\subsection{Visual Feature Extraction}
Considering the complexity of 3D CT-scan images, we adopted a recently introduced CT-CLIP \cite{hamamci2024foundation} image encoder that is pre-trained in a contrastive setting using a large number of 3D CT-focused image-text pairs. This encoder is based on the CT-ViT \cite{hamamci2024generatect} architecture that processes the input $\mathbf{x}\in \mathbb{R}^{(240)\times 480 \times 480}$ as non-overlapping patches of shape $(10)\times 20 \times 20$, where $10$ is the temporal patch size $t_p$ and $20$ denotes spatial patch sizes $p_1$ and $p_2$. These patches are reshaped to $B\times T\times H\times W\times (C . t_p . p_1 . p_2)$, where $B$ is the batch size and $T$ is the temporal patch count. A linear transformation is then applied to obtain frame embeddings of shape $B\times T\times \frac{H}{p_1}\times \frac{W}{p_2}\times v$, where $v$ is the final required latent representation dimension. The spatial transformer processes this reshaped tensor, maintaining the same size. Subsequently, it passes through a causal transformer, yielding a tensor of shape $(\frac{H}{p_1} . \frac{W}{p_2})\times (B . T)\times v$. Such a combination of spatial and causal transformations ensures the retention of 3D information throughout the network. Finally, to obtain the latent representation, the embeddings obtained are processed through a linear layer to convert them into a vector of dimension $v$, where $v$ is set to 512. This latent vector is used as input for the projection network. For in-domain data samples, we freeze all parameters of $f_e$, whereas for out-of-domain data samples, we fine-tuned the last layer. 

\subsection{Anomaly-Aware Prefix Projection Network}
Our proposed model's main trainable component is a lightweight, transformer-based projection network, $f_m$. Its purpose is to align the image-derived embeddings with the LLM's token space. Unlike traditional projectors that only use visual features as a single token or sequence of tokens, our network is \textbf{anomaly-aware} that generates a semantically disentangled sequence of visual and clinical context and maps them into a sequence of LLM-style tokens. It leverages both a latent vector from the image encoder and a set of multi-label diagnostic classification logits to create a more comprehensive input. This design is motivated by the need to explicitly provide the model with high-level clinical priors, thus retaining critical diagnostic cues that might otherwise be lost.


To integrate the diagnostic information, we concatenate the multi-anomaly classification logits, $\mathbf{l}_i$, with the image features $\mathbf{z}_i$ obtained from $f_e$. This fusion creates a rich, combined representation, and can be defined as $\hat{\mathbf{z}}_i = [\mathbf{z}_i \oplus \mathbf{l}_i]\in \mathbb{R}^{v+c}$, where $\oplus$ denotes concatenation. Initially, these fused embeddings $\hat{\mathbf{z}}_i$ are projected into a structured sequence using a linear transformation. The obtained structured embeddings and a learnable constant serve as input for the projection network. The learnable constant extracts relevant information from the embeddings and adjusts the network to new data samples.  The transformer layers within our projection network then operate in a self-attention setting to capture complex dependencies and optimize the representation. The output of this process is the final prefix, a fixed-length embedding sequence,  called \textit{anomaly-aware prefix}, which is then concatenated with the textual report embeddings 
$\hat{\mathbf{R}}_i=\{\hat{\mathbf{r}}_1,\hat{\mathbf{r}}_2,...,\hat{\mathbf{r}}_N\}$, $\hat{\mathbf{r}}_j\in \mathbb{R}^h$,
before being fed to the frozen LLM $f_d$. Here $h$ represents the LLM's hidden size. This design ensures that the model is conditioned not only on the raw visual content but also on clinically significant multi-abnormality markers, helping to generate more factual and diagnostically precise reports. The classification logits are kept soft and frozen during report generation to provide valuable clinical priors without introducing backpropagation errors into the classifier. This strategy conditions the LLM to generate clinically accurate reports. 

\textbf{Training Objective:} The training objective is to optimize the lightweight, trainable projection network, $f_m$, while the image encoder $f_e$ and text decoder $f_d$ remain frozen. This objective is achieved by minimizing the negative log-likelihood of the target report sequence. The loss function is defined as: 
\begin{equation}
    \label{eq:loss}
    \mathcal{L} = - \sum^{M}_{i=1}\sum^{N}_{j=1} log\hspace{0.2em}p_\theta(\hat{\mathbf{r}}_{i,j}|\hat{\mathbf{z}}_i, \hat{\mathbf{r}}_{i,1}, ..., \hat{\mathbf{r}}_{i,{j-1}}).  
\end{equation}
Here, $M$ is the number of reports in the dataset, and $N$ is the number of tokens in the $i$-th report. The term $p_\theta(\hat{\mathbf{r}}_{i,j}|.)$ represents the probability of predicting the $j$-th token of the $i$-th report, $r_{i,j}$, which is conditioned on the entire prefix, $\hat{\mathbf{z}}_i$ and all previously generated tokens.  The parameters being optimized, denoted by $\theta$ belong exclusively to the projection network, $f_m$. The goal is to train this network to generate a prefix representation,  $\hat{\mathbf{z}}_i$, that effectively conditions the frozen LLM ($f_d$) to produce the correct report tokens, $r_{i,j}$, in an autoregressive manner. This approach ensures that the LLM's vast, pre-trained knowledge is leveraged while the model learns to bridge the modality gap using only a minimal number of trainable parameters.

During the training phase, the projection network is optimized using the training loss $\mathcal{L}$ (Eq.\ref{eq:loss}), constrained by the concatenated prefix and report embeddings ($\hat{\mathbf{z}}_i, \hat{\mathbf{r}}_{i,1}, ..., \hat{\mathbf{r}}_{i,{j-1}}$). In contrast, only the prefix projections ($\hat{\mathbf{z}}_i$) are used during evaluation, and the model generates the report tokens iteratively in an autoregressive manner, selecting each token based on the highest computed probabilities in the process. A detailed algorithm for projection network training and the classification module are discussed in the supplementary material. 




\subsection{Report Decoder}
Report decoder $f_d$ is a pre-trained LLM, extensively trained on a large corpus of generic text data. We adopted the LLaMA-3.2-1B\footnote{https://huggingface.co/meta-llama/Llama-3.2-1B} model, originally designed to generate relevant text responses when given an input text prompt. This limits its ability to handle visual content, as they are encoded in a different format. Moreover, fine-tuning such large models could interfere with the model's generalizability and lead to suboptimal performance. Therefore, we used the decoder's pre-trained weights without fine-tuning to leverage its already captured rich and hierarchical text representations and avoid overfitting on small medical datasets. However, to align the medical semantic features and the textual data in the report, we used the projection network that reformulates visual cues and the semantics of the reports in a manner that LLaMA-3.2-1B can effectively interpret and process. The decoder $f_d$ iteratively generates the report tokens when given some cues in the form of prefix projections. This process can be defined as:

\begin{equation}
P(w_t \mid w_{<t}, \hat{\mathbf{z}}_{i,1:L_p}) = \text{softmax}(w_{{out}} \cdot f_d(w_{<t}, \hat{\mathbf{z}}_{i,1:L_p}),
\end{equation}

where $L_p$ is the prefix length, $w_{<t}$ signifies the previously generated tokens and $\mathbf{w}_{out}$ is the output token projection matrix that maps LLaMA's final hidden state to vocabulary space.

\section{Experiments}
\subsection{Configurations}
\textbf{Datasets:} We conducted experiments using two publicly available large datasets: CT-RATE \cite{hamamci2024foundation} and INSPECT \cite{huang2023inspect} datasets. CT-RATE is used for in-domain evaluation since the CT-CLIP~\cite{hamamci2024foundation} encoder is pre-trained on it, while INSPECT serves as an out-of-domain dataset. CT-RATE consists of 50,188 non-contrast chest CT volumes with 18 multi-abnormality labels. We used its official split, i.e., 47,149 and 3,039 scans for training and testing, respectively. The findings section is used for report generation.




\begin{table}[t]
\centering
\resizebox{0.8\linewidth}{!}{
\begin{tabular}{c|c|ccccc}
\begin{tabular}[c]{@{}c@{}}Model Name\end{tabular}                   & Method         & \begin{tabular}[c]{@{}c@{}}Avg. BLEU (1-4)\end{tabular} & METEOR & \begin{tabular}[c]{@{}c@{}}Avg. ROUGE       (1-2)\end{tabular} & \begin{tabular}[c]{@{}c@{}}ROUGE-L\end{tabular} & \begin{tabular}[c]{@{}c@{}}BERTScore-F1\end{tabular}\\ 
\midrule 
 \multirow{2}{*}{DistilGPT2}   & Baseline\textsuperscript{\ref{clip-text-decoder}} & 0.2660                                                         & 0.4205 & 0.4416                                                             & 0.4166  & 0.8832             \\                            & V-1  & \textbf{0.2925}                                                        & \textbf{0.4271} & \textbf{0.4501 }                                                            & \textbf{0.4281}  & \textbf{0.8925}                                                          \\
                          \multirow{2}{*}{GPT2}         & Baseline\textsuperscript{\ref{clip-text-decoder}} & 0.2476                                                        & 0.4013 & 0.4380                                                              & 0.4014  & 0.8810                                                            \\                               & V-1  & \textbf{0.3065 }                                                       & \textbf{0.4366} & \textbf{0.4684 }                                                            & \textbf{0.4315}  & \textbf{0.8932}                                                           \\
                          GPT2- & Baseline\textsuperscript{\ref{clip-text-decoder}} & 0.2658                                                        & 0.4321 & \textbf{0.4676}                                                             & \textbf{0.4314}  & 0.8858                                                           \\ 
                           Medium   & V-1  & \textbf{0.3037}                                                        & \textbf{0.4351} & 0.4672                                                             & 0.4313  & \textbf{0.8931}                                                          \\ 
                         LLaMA- & Baseline\textsuperscript{\ref{clip-text-decoder}} & 0.2637                                                        & 0.4335 & \textbf{0.4677}                                                             & \textbf{0.4363}  & 0.8864                                                            \\
                            3.2-1B & V-1  & \textbf{0.3427}                                                        & \textbf{0.4467} & 0.4638                                                             & 0.4062  & \textbf{0.8891 }                                                            \\ \hline         
\end{tabular}
} \caption{Comparative analysis of baseline and variant V-1. Best values are in \textbf{bold}.}
\label{tab:v1}
\end{table}
\begin{figure}[t]
    \centering
    \includegraphics[width=\columnwidth]{ 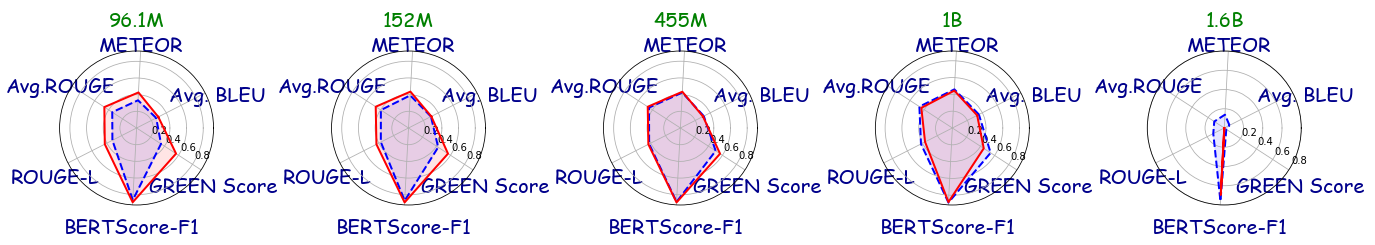}
    \caption{Radar plots showing the impact of fine-tuning (solid) and freezing (dashed) parameters of variable-sized LLMs (96.1M-1B).  Sequence: DistilGPT2, GPT2, GPT2-medium, LLaMA-3.2-1B, and BioGPT-Large. }
    \label{fig:radar-plot}
\end{figure}



INSPECT comprises CT-scans acquired from 19,402 patients, focused on pulmonary embolism. After removing redundant data, we used 17,730 and 3,506 scans for training and testing, respectively. Due to the unavailability of the official split details, a stratified sampling was performed based on 21 distinct anomalies. 
More details on dataset split preparation are in the supplementary material.


\noindent \textbf{Pre-processing and Training Details:} Inspired by CT-CLIP \cite{hamamci2024foundation}, each CT volume is resized for a uniform spacing, i.e., 0.75 mm in the x-axis and the y-axis, and 1.5 mm in the z-axis. All experiments are conducted using $480\times 480\times 240$ resolution with Hounsfield Units (H.U.) clipped to -1000 to 1000 range, followed by normalization. The experiments are performed on NVIDIA-A100 GPU using PyTorch framework, training all models for 10 epochs to ensure a fair comparison. The projection module is trained using an Adam optimizer with a learning rate of 2e-5. 

\noindent \textbf{Baseline:} We compared our method with various state-of-the-art techniques in terms of (a) vision-text alignment approach, (b) 3D medical image to report generation task, and (c) different sized LLMs adoption. We primarily compare our work with R2GenGPT \cite{wang2023r2gengpt} due to its overall architectural similarities and emphasis on medical report generation. Although it was originally proposed for 2D images, we performed its training by feeding it our extracted visual latent embeddings. R2GenGPT uses a frozen LLM and, hence, to additionally compare with trainable LLM, we created our baseline using basic clip-to-text decoder\footnote{\label{clip-text-decoder}https://github.com/fkodom/clip-text-decoder} architecture, using same visual embeddings but a conventional projection approach and fine-tunes the LLM during training. 

We compared our proposed model with R2GenGPT using variable LLMs: domain-specialized LLM (BioGPT-Large \cite{10.1093/bib/bbac409}), an LLM with the same number of parameters (LLaMA-3.2-1B-Instruct), and an LLM with higher parameters originally adopted in their study (LLaMA-2-7b-chat-hf). The other baseline is additionally used to validate our three proposed alignment module variants (\textit{V-1}, \textit{V2}, and \textit{V-3}, shown in Fig. \ref{fig:variants}). Further, we performed a comparative analysis with recent work in the same domain. 3D CT report generation is understudied, with only a few methods available but without publicly available code for reproducibility (E3D-GPT, CT-ARG). Hence, we relied on reported values only. These works are closest in scope, along with CT2Rep.



\begin{figure*}[t]
    \centering
    \includegraphics[width=0.85\linewidth]{ 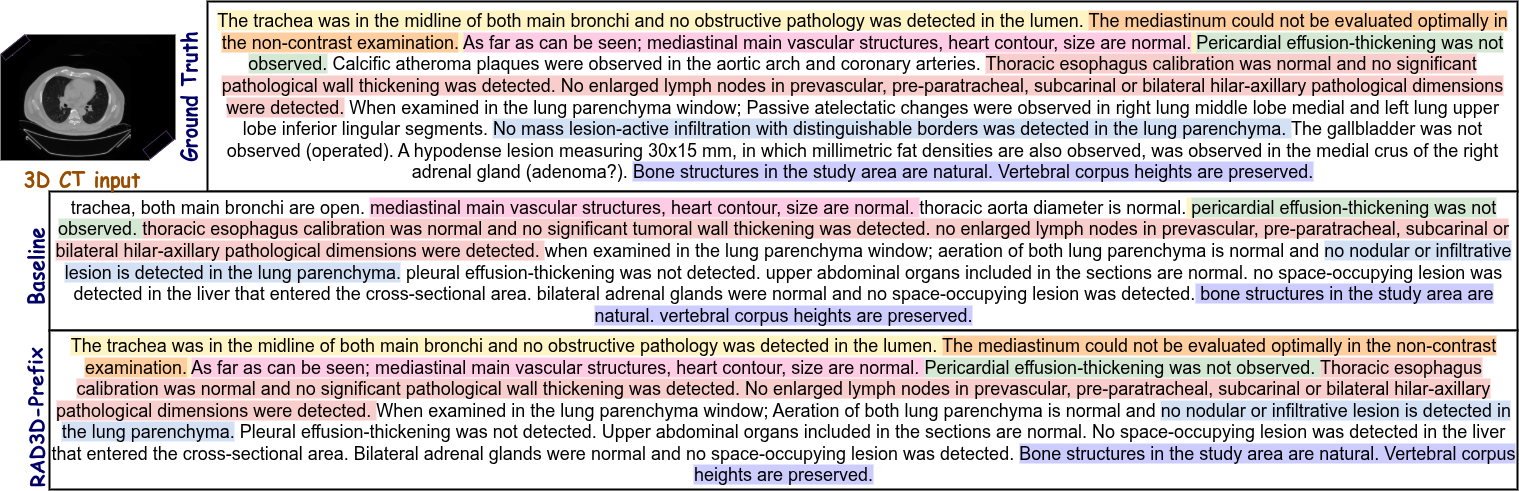}
    \caption{Qualitative example of the baseline and RAD3D-Prefix. Matching sentence pairs are highlighted in the same color.}
    \label{fig:qual}
\end{figure*}

\subsection{Results}
This section compares the baseline approach and three proposed variants.

\noindent \textbf{Baseline vs. Variant \textit{V-1}:} To assess the significance of including visual embedding as a prefix (Variant \textit{V-1}), we compared its report generation outcomes with that obtained using baseline (without prefix). As shown in Table \ref{tab:v1}, the majority of metrics favour the variant \textit{V-1} across four different LLMs with varying parameter count. The overall results indicate an increase of 9.96\% to 29.95\% in Avg. BLEU, 0.69\% to 8.80\% in METEOR, and a maximum of 6.94\% and 7.50\% improvement in Avg. ROUGE (1-2) and ROUGE-L, respectively. Similarly, an increase of about 1.39\% is observed in BERTScore-F1. The reported improvements highlight the superior performance of variant \textit{V-1} over the baseline, underscoring the importance of the visual prefix in enhancing report generation.

\noindent \textbf{Variant \textit{V-1} vs. Variant \textit{V-2}:} This comparison discusses the key concept of this work, i.e., the impact of fine-tuning and freezing parameters of smaller ($<1B$) and larger ($\approx 1B+$) LLMs. Fig. \ref{fig:radar-plot} clearly illustrates that models with parameter sizes ranging from 96.1M to 455M exhibit a performance decline across various metrics, including the GREEN Score, when adopted as frozen. 
Note that both fine-tuned (variant \textit{V-1}) and frozen (variant \textit{V-2}) models with 96.1M to 1B parameters utilize the same projection module, involving visual embeddings. In contrast, the last model, BioGPT-Large, is trained using the standard baseline approach for fine-tuning, whereas the R2GenGPT model deploys it as a frozen module. The model exhibited severe overfitting when fine-tuned.

We can conclude two findings from this comparison: (a) With the increase in parameter number, the performance gap between the frozen and fine-tuned models starts to decrease. Further increasing the parameters (beyond 1B) improved the frozen model's performance compared to its fine-tuned counterpart. (b) Our proposed projection network outperforms the conventional mapping criteria followed in the existing work, as demonstrated by the degraded performance of the baseline models in the last graph plot. Although the overall trend between the frozen and fine-tuned models remains consistent, a clear distinction in performance is evident. We further compare the \textit{V-3} variant in the ablation study section below. Also, qualitative results are shown in Fig. \ref{fig:qual}. More qualitative samples are given in the supplementary material.

\begin{table*}[t]
\centering
\resizebox{\linewidth}{!}{
\begin{tabular}{c|c|c|cccccc}
 Method   & LLM                   & \begin{tabular}[c]{@{}c@{}}No. of \\ Parameters\\ in LLM\end{tabular} & \begin{tabular}[c]{@{}c@{}}Avg. BLEU\\ (1-4)\end{tabular} & METEOR & \begin{tabular}[c]{@{}c@{}}Avg. ROUGE\\ (1-2)\end{tabular} & ROUGE-L & BERTScore-F1 & \begin{tabular}[c]{@{}c@{}} GREEN\\ Score \\ (Clinical Efficacy)\end{tabular} \\ \midrule
 Baseline\textsuperscript{\ref{clip-text-decoder}} & BioGPT-Large          & 1.6B                                                                                & -         & -      & -                                                          & -       & 0.6932       & -                                                                   \\
                         
                          R2GenGPT  & BioGPT-Large          & 1.6B                                                                                & 0.0611    & 0.1385 & 0.1241                                                     & 0.1284  & 0.7622       & 0.0254                                                              \\
                          R2GenGPT & LLaMA-3.2-1B-Instruct & 1B                                                                                  & 0.2902    & 0.3762 & 0.39485                                                    & 0.3468  & 0.8751       & 0.4120                                                                    \\
                          R2GenGPT & LLaMA-2-7b-chat-hf          & 7B                                                                                & 0.3523    & 0.4509 & 0.4640                                                     & 0.4038  & 0.8886       &   0.5041                                                            \\
                          
                         \textbf{RAD3D-Prefix}     & LLaMA-3.2-1B          & 1B                                                                                  & \textbf{0.3637}    & \textbf{0.4694} & 0.4256                                                     & 0.4190   & 0.8883       & \textbf{0.5488}                                                              \\  \textbf{RAD3D-Prefix}     & DeepSeek-R1-Distill-LLaMA-8B         & 8B                                                                                  &   
                         0.3405                                                    & 0.4459 & \textbf{0.4704}                                                     & \textbf{0.4339}  & \textbf{0.8909 }       & 0.5443 \\
                         \hline
                                            
\end{tabular}
}
\caption{Comparison with the state-of-the-art approach of vision and text embedding alignment and different sized LLM training. The analysis is divided into three categories: (a) LLM (both fine-tuned: baseline, row 1 and frozen: row 2) with specialized domain pre-training, (b) frozen LLM with the same parameter count (row 3) and (c) with higher parameter count using a conventional alignment approach (row 4). Values in \textbf{bold} denote the best outcome.}
\label{tab:sota-1}
\end{table*}


\noindent \textbf{Metrics:} We report results on four widely used NLG metrics, namely, BLEU (1-4)~\cite{papineni-etal-2002-bleu}, METEOR~\cite{banerjee-lavie-2005-meteor}, ROUGE (1, 2, L)~\cite{lin-2004-rouge} and BERTScore-F1~\cite{zhangbertscore}. In addition, we adopt the GREEN score~\cite{ostmeier-etal-2024-green}, which is specifically relevant to the medical domain. BLEU measures n-gram precision, with a penalty for short translations, whereas METEOR incorporates synonym matching and word order penalties. ROUGE focuses on n-gram recall, and the BertScore-F1 focuses on semantic similarity using a language model. The GREEN score focuses on the factual correctness of clinical information.


\begin{table}[b]
\centering
\resizebox{0.6\columnwidth}{!}{
\begin{tabular}{c|c|ccccccc}
Dataset & Method               & BLEU-4 & METEOR & ROUGE-1 & BERTScore-F1 & GREEN \\ \midrule 
\multirow{4}{*}{CT-RATE} &E3D-GPT\textsuperscript{\textdagger}                 & -      & 0.4179 & 0.5260       & 0.8797 &  -    \\
& CT-AGRG\textsuperscript{\textdagger}     & 0.1720  & 0.1960  & -      & 0.8670 & -       \\
& CT2Rep      & \textbf{0.3212}      & 0.4543  & \textbf{0.5899}       & \textbf{0.8929}  & 0.5247      \\
& RAD3D-Prefix                   & 0.2779 & \textbf{0.4694} & 0.5780  & 0.8894 &  \textbf{0.5488}   \\
\hline
\multirow{2}{*}{INSPECT} & CT2Rep        & 5.74e-06       & 0.1207          &      0.1892                                                                                                   &   0.8629   &  0.2219   \\
& RAD3D-Prefix                     & \textbf{0.0344} & \textbf{0.2122} & \textbf{0.2268}   & \textbf{0.8670}  & \textbf{0.2400}   \\
\end{tabular}
}\caption{Comparison of state-of-the-art approach for 3D images to clinical report generation. \textsuperscript{\textdagger}The results are from the original paper on the same test split. - denotes the unreported metrics.}
\label{tab: sota-2}
\end{table}

\subsection{Comparison with State-of-the-art Methods}
In this section, we compare our approach with state-of-the-art methods for medical report generation. We focus on models with similar (i) architectural frameworks (R2GenGPT) and (ii) methodological objectives (3D image to report generation) to ensure a meaningful comparison. Table \ref{tab:sota-1} presents a comparative analysis of our approach against the first category models with (a) specialized domain LLM, (b) same parameter count LLM, and (c) higher parameter count LLM. We used BioGPT-Large \cite{10.1093/bib/bbac409} as a specialized domain LLM in our baseline setting, i.e., a conventional mapping network with fine-tuning. The model experiences significant overfitting, and hence, the results are not reported in the table. We further used the same model with R2GenGPT, i.e., as a frozen module and demonstrated better performance than its baseline version. 

We further replaced LLaMA-3.2-1B with a larger LLM, DeepSeek-R1-Distill-LLaMA-8B. It can be observed that across different LLM configurations, RAD3D-Prefix consistently outperforms the R2GenGPT baselines. Using LLaMA-3.2-1B, RAD3D-Prefix achieves the highest Avg. BLEU, METEOR, and the best GREEN score (0.5488), indicating improved clinical relevance. When paired with the larger DeepSeek-R1-Distill-LLaMA-8B, RAD3D-Prefix further improves ROUGE metrics and BERTScore-F1, showing that the framework effectively scales with stronger LLMs. Although R2GenGPT performs better with a 7B LLM compared to its 1B version, RAD3D-Prefix with only a 1B LLM already surpasses or matches the 7B baseline across most metrics, highlighting the efficiency and scalability of the proposed approach.

In addition to the specialized LLMs, we trained R2GenGPT with two more LLMs, LLaMA-3.2-1B-Instruct and LLaMA-2-7b-chat-hf. We selected LLaMA-3.2-1B-Instruct to conduct a fair comparison with the same version and size of LLaMA. An instruction variant of LLaMA was selected to serve a similar purpose to the original LLM (LLaMA-2-7b-chat-hf) used by R2GenGPT. The results indicate that our model clearly outperforms the same-sized LLM, i.e., LLaMA-3.2-1B-Instruct, achieving a GREEN score of 0.5488. Furthermore, even when a larger LLM is deployed in a similar setting, our model outperformed it on several metrics and performed comparably on others. For a fair comparison, we also replaced our transformer-based projection network with a linear layer (similar to R2GenGPT) for which details are given in the supplementary material.  

We further evaluated our approach against existing 3D medical image-to-report generation methods. Table \ref{tab: sota-2} presents results from several recent techniques applied to the same dataset. Our method, \textbf{RAD3D-Prefix}, outperforms most existing techniques on the CT-RATE dataset, with the exception of CT2Rep on certain NLG metrics. However, our model's superiority is evident in its clinical relevance, as demonstrated by a higher GREEN score compared to CT2Rep. Furthermore, \textbf{RAD3D-Prefix} significantly outperformed CT2Rep on the out-of-domain INSPECT dataset, underscoring its generalizability and robustness in real-world scenarios.


\begin{table*}[t]
\centering
\resizebox{\linewidth}{!}{
\begin{tabular}{c|c|ccc|ccccccc}
Dataset                  & Variant & \begin{tabular}[c]{@{}c@{}}Visual \\Prefix \end{tabular} & \begin{tabular}[c]{@{}c@{}}Frozen \\ LLM \end{tabular} & \begin{tabular}[c]{@{}c@{}}Multi-label \\ Classification \\ Logits\end{tabular} & \begin{tabular}[c]{@{}c@{}}Avg. BLEU\\ (1-4)\end{tabular} & METEOR & \begin{tabular}[c]{@{}c@{}}Avg. ROUGE\\ (1-2)\end{tabular} & \begin{tabular}[c]{@{}c@{}} ROUGE- \\ L \end{tabular}& \begin{tabular}[c]{@{}c@{}}BERTScore- \\F1 \end{tabular} & \begin{tabular}[c]{@{}c@{}}GREEN \\ (Clinical \\ Efficacy)\end{tabular} & \begin{tabular}[c]{@{}c@{}}No. of \\ Trainable \\ Parameters.\end{tabular} \\ \midrule
\multirow{4}{*}{CT-RATE} & V-1 &        \ding{51}          &   \ding{55}         &      \ding{55}                                                                           & 0.3315                                                    & 0.4456 & 0.4402                                                    & 0.3646 & 0.8826       & 0.4454 & 1.51B                                                                      \\
                    & V-2     &     \ding{51}           & \ding{51}            &  \ding{55}                                                                               & 0.3543                                                    & 0.4604 & 0.4677                                                     & 0.4190   & 0.8883       & 0.5428   & 279.09M                                                                    \\
                   & V-3      &      \ding{51}          & \ding{51}             &   \ding{51}                                                                               & \textbf{0.3637 }                                                   & \textbf{0.4694} & \textbf{0.4716}                                                     & \textbf{0.4256}  & \textbf{0.8894}       & \textbf{0.5488} & 279.46M                                                                      \\ \hline
\multirow{4}{*}{INSPECT} 
                     & V-1    &     \ding{51}           &  \ding{55}           &  \ding{55}                                                                                & \textbf{0.0984 }                                                   & \textbf{0.2473} & 0.1586                                                     & 0.1747  & 0.8505       &  0.1565 & 1.51B                                                                           \\
                      & V-2   &   \ding{51}             & \ding{51}             & \ding{55}                                                                                  & 0.0579                                                    & 0.2028 & 0.1810                                                      & 0.2189  & 0.8659       & 0.2355  & 279.09M                                                                     \\
                      & V-3   &     \ding{51}           & \ding{51}            &   \ding{51}                                                                               & 0.0657                                                    & 0.2122 & \textbf{0.1889}                                                     & \textbf{0.2268}  & \textbf{0.8670}        & \textbf{0.2400} & 279.46M                                                                     
\end{tabular}
}
\caption{Comparative analysis of the role of presence/absence of different features (Visual Prefix, Frozen LLM, and Multi-label Classification Logits) in the proposed RAD3D-Prefix model.}
\label{tab: ablation}
\end{table*}

\subsection{Ablation Study and Out-of-domain Performance}\label{sec:ablation}
We performed an ablation study to assess the significance of different concepts introduced in \textbf{RAD3D-Prefix}, especially the anomaly-aware projection module, as shown in Table \ref{tab: ablation}. Additionally, INSPECT, an out-of-domain dataset is utilized to assess the impact of various features when applied to a dataset from a different distribution. It is evident that incorporating each feature leads to improvements across most metrics. The GREEN score, which is considered the most relevant metric from a clinical perspective, demonstrates a 23.22\% improvement in the variant \textit{V-3} setting compared to variant \textit{V-1}. Similarly, using the INSPECT dataset, a 53.36\% improvement is observed.

\begin{figure*}[t]
    \centering

        \includegraphics[width=\linewidth]{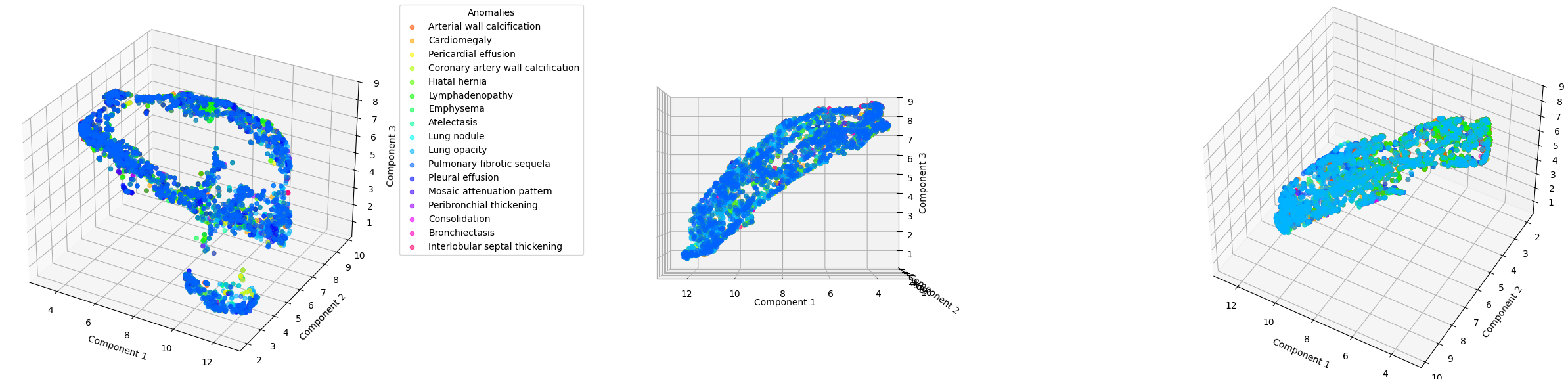}
        \label{fig:umap_v2}
    \hfill
        \includegraphics[width=\linewidth]{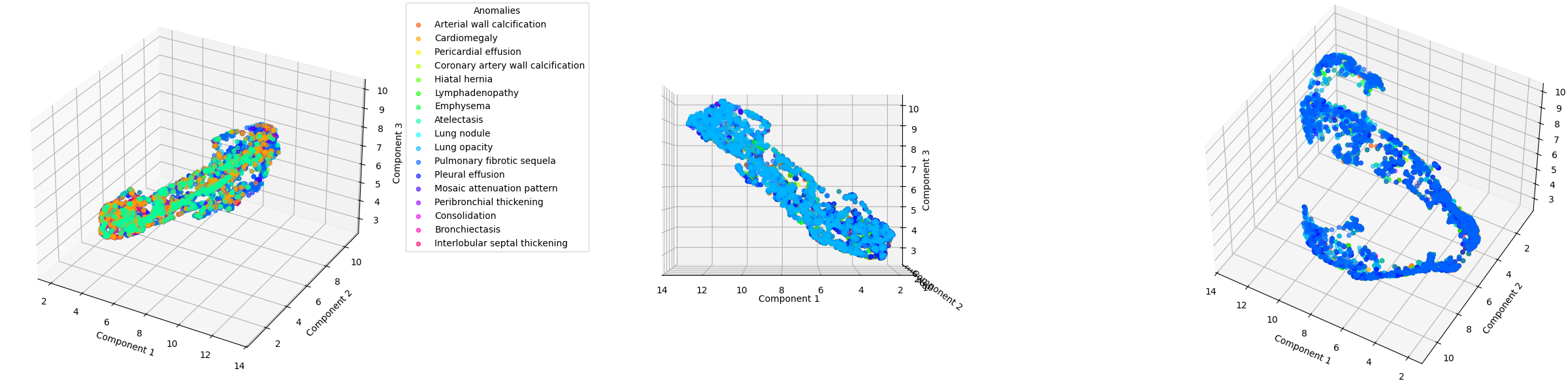}
        \label{fig:umap_v3}
\caption{UMAP visualizations across different projection networks: V-2 (top) and V-3 (bottom) .}
    \label{fig:umap}
\end{figure*}

\noindent \textbf{UMAP Visualization:} To inspect the learned representation in the \textit{V-2} and \textit{V-3} variants, we projected the embeddings onto the three components and visualized them from multiple viewpoints. In Fig. \ref{fig:umap} (\textit{V-2}), the blue/cyan ribbon is looser, less structured, and appears more scattered, suggesting noisy embedding. 
In Fig. \ref{fig:umap} (\textit{V-3}), the blue/cyan ribbon is tighter and more clearly defined. It yields a more compact S‑shaped ribbon, reflecting a more structured organization of the latent space while preserving the overall disease continuum. 
It has a denser core with fewer isolated ``outlier'' points, 
demonstrating that the anomaly-aware mechanism creates unique vectors for structurally important diseases.

\subsection{Classifier Results and Dependency Analysis}
\label{classification}
\begin{figure*}[t]
    \centering
    \includegraphics[width=0.49\linewidth]{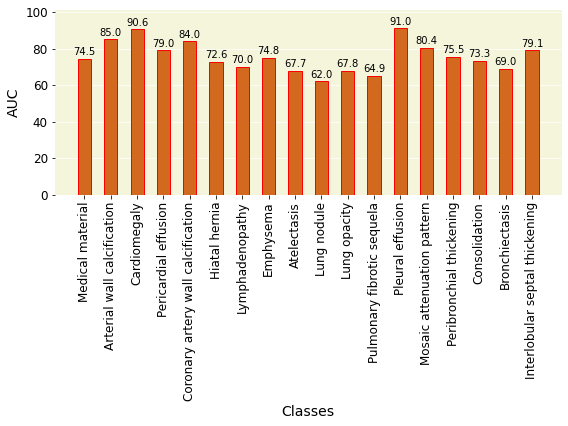}
    \hfill \includegraphics[width=0.49\linewidth]{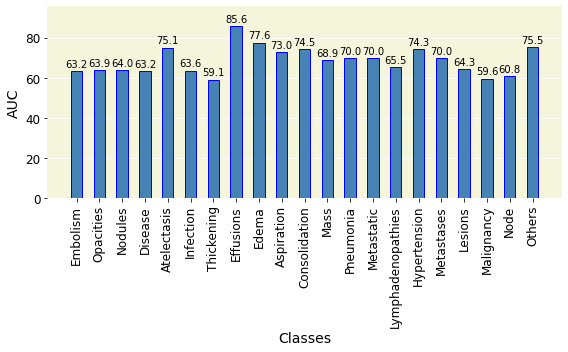}
    \caption{Classification results on 18 and 21 multi-abnormality labels of the CT-RATE (left) and INSPECT (right) dataset, respectively.}
    \label{fig:classification}
\end{figure*}

\textbf{Results:} To obtain the multi-abnormality classification logits, we fine-tuned the last layer of the CT-CLIP encoder and added a classification layer. (Fig. \ref{fig:overalframework}). Note that the same CT-CLIP encoder used for CT scan feature extraction is employed for this purpose; however, as an independent module, separate from the report generation pipeline. For the CT-RATE dataset, we used its original weights from \cite{hamamci2024foundation} since the CT-CLIP encoder is pre-trained with CT-RATE and for the INSPECT dataset, its last layer is fine-tuned. The training process involves binary cross-entropy loss with class-wise weights to address class imbalance. A multi-label classification is performed featuring the 18 and 21 abnormalities of the CT-RATE and INSPECT datasets, respectively. Therefore, the output dimension equals the number of findings (CT-RATE: 18, INSPECT:21). We concatenate these 1D logits with image embeddings, and this adds anomaly-specific priors to guide generation. The classification results are shown in Fig. \ref{fig:classification}. These results demonstrate strong anomaly detection, capturing clinical patterns that enhance report generation, particularly for critical conditions (e.g., effusion: 91.0\%, cardiomegaly: 90.6\%).\\

\noindent \textbf{Dependency Analysis:} We observed multiple cases where the anomaly classifier assigned relatively high confidence to a finding, yet the report generation system correctly described the finding in agreement with the ground-truth report. Some sample reports from three different reports are shown in Table \ref{error}. Notably, the classifier assigned relatively high confidence scores (>0.6 and >0.5) to pericardial effusion and hiatal hernia, respectively. Despite these misleading classifier signals, the generated reports correctly stated that these abnormalities were not observed, consistent with the reference reports. Similarly, a basic \textit{Hiatal hernia} prior is refined into a specific subtype, \textit{``Sliding type hiatal hernia...''}, which does not exist in training labels. These examples demonstrate that the report generator is not solely based on classifier predictions and can override incorrect classifier signals or refine clinical details using information retained in the visual features.

\begin{table}[t]
\centering
\caption{Samples showing visual tokens overriding classifier errors and adding subtype.}
\label{tab:overrides_and_refinement}
\resizebox{\textwidth}{!}{%
\begin{tabular}{@{}lllll@{}}
\toprule
\textbf{Mechanism Type} & \textbf{Target Condition} & \textbf{Classifier Logit} & \textbf{Ground Truth Report} & \textbf{RAD3D-Prefix Output} \\ \midrule
\multirow{3}{*}{\textbf{False Positive Override}} & Pericardial Effusion & 0.62017 & ``Pericardial effusion-thickening not observed.'' & ``Pericardial effusion-thickening was not observed.'' \\
 & Pericardial Effusion & 0.63840 & ``Pericardial effusion-thickening not observed.'' & ``Pericardial effusion-thickening was not observed.'' \\
 & Hiatal Hernia & 0.51572 & Pathology absent in official scan study. & Pathology accurately suppressed from report text. \\ \midrule
\textbf{Subtype Addition} & Hiatal Hernia & 0.52510 & \textbf{``Sliding type hiatal hernia} was observed...'' & \textbf{``Sliding type hiatal hernia} was observed...'' \\ \bottomrule
\end{tabular}%
}
\label{error}
\end{table}

\subsection{Statistical Significance Evaluation and Reader Study}
\begin{figure}[t]
    \centering
    \includegraphics[width=0.7\linewidth]{ 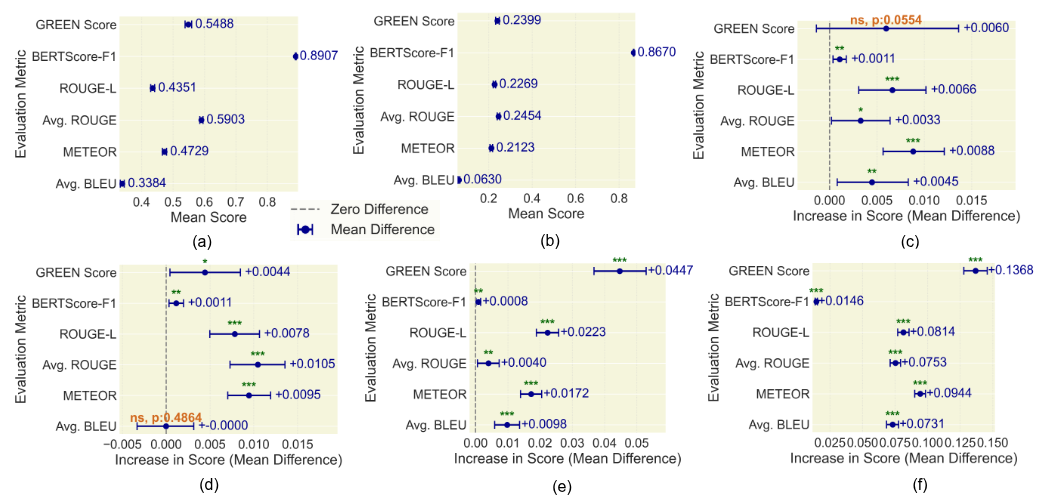}
    \caption{Forest plots of mean differences (95\% CIs) for RAD3D-Prefix on (a) CT-RATE and (b) INSPECT, with comparisons to (c–f) Variant V2, R2GenGPT (LLaMA-2-7b-chat-hf), and RGenGPT (LLaMA-3.2-1B-Instruct). Asterisks denote significance levels: $^\ast p<0.05$, $^{\ast\ast} p<0.01$, $^{\ast\ast\ast} p<0.001$; ns indicates $p \ge 0.05$.  }
    \label{fig:stat}
\end{figure}

\textbf{Statistical Significance Evaluation:} 
To evaluate the significance of improvements, we performed bootstrapping with 5,000 iterations. 
Fig.~\ref{fig:stat} (a) and Fig.~\ref{fig:stat} (b) demonstrate our model's consistent and stable performance, as indicated by the narrow bootstrapped 95\% CI across all metrics using the CT-RATE and INSPECT datasets, respectively. In addition, we compared \textit{V-2} and \textit{V-3} variants across the two datasets (Fig.~\ref{fig:stat} (c) and (d)) to assess the significance of introducing classification logits into our proposed projection network. We observed that five out of six metrics illustrate statistically significant improvements ($p < 0.05$), signifying robust and generalized efficacy of the \textbf{RAD3D-Prefix} that hold across independent datasets and evaluation dimensions, thereby validating the significance of the anomaly-aware prefix. In Fig.~\ref{fig:stat} (c), although the GREEN score improvement does not meet the significance threshold but the marginal p-value ($p = 0.0554$) suggests that the improvement is borderline meaningful and may warrant further investigation. Similarly, in Fig.~\ref{fig:stat} (d), the mean difference in Avg. BLEU is minimally negative ($\text{-0.000031}$), suggesting that \textit{V-2} achieved a marginally higher score. However, since the $p$-value is high ($p=0.4864$), this difference is not statistically significant. We further compared our model with the baseline R2GenGPT with a larger (Fig.~\ref{fig:stat} (e)) and similar-sized LLM (Fig.~\ref{fig:stat} (f)). The consistent performance gain with all metrics meeting the significance criterion ($p < 0.05$) highlights the robustness of our method. 

\noindent \textbf{Reader Study:} We conducted a reader study with two clinical experts who evaluated 100 randomly selected predicted reports from the baseline and two variants \textit{V-2} and \textit{V-3}. Averaging scores from both clinicians demonstrates that V-3 achieved the highest clinically relevant outcomes, improving scores by 9.8\% over the baseline and 3.7\% over V-2. More details are in the supplementary material.  

%


\section{Conclusion}
In this paper, we present RAD3D-Prefix model, with a \textit{lightweight anomaly-aware prefix conditioning} module for the generation of 3D CT reports. This approach effectively bridges the semantic clinical gap by aligning high-dimensional volumetric features and structured diagnostic priors with the frozen LLM's text embedding space. Despite requiring fewer trainable parameters, the proposed model achieves a 4.6\% improvement in GREEN score over the SoTA. Additionally, we compare different variants to highlight scenarios where one variant can be preferred compared to another variant. Through a systematic study, we investigate the impact of model scale and adaptation strategy, establishing practical recommendations for choosing between freezing and fine-tuning paradigms in 3D medical imaging.

\section*{Appendix}
\section*{Dataset Details}
\label{dataset_details}
\subsubsection*{CT-RATE}
The CT-RATE dataset consists of 25,692 non-contrast chest CT volumes, which are expanded to 50,188 using various reconstruction methods. These scans come from 21,304 unique patients and are accompanied by corresponding radiology text reports, 18 multi-abnormality labels, and metadata. During training, samples from 20,000 patients (47,149 scans) are used, while the remaining samples from 1,304 patients (3,039 scans) are reserved for testing. The associated radiology reports are segmented into four sections: clinical information, technique, findings, and impression. However, only the findings section is used to train the report generation model. For the classification task, the publicly available 18 multi-abnormality labels are used.

 \begin{table}[!]
\centering
\begin{tabular}{lll}
\textbf{Anomaly}  & \textbf{Training Set} & \textbf{Test Set} \\
\toprule 
Embolism          & 3691                  & 739               \\
Opacities         & 3788                  & 758               \\
Nodules           & 4296                  & 859               \\
Disease           & 2592                  & 518               \\
Atelectasis       & 2916                  & 583               \\
Infection         & 2744                  & 549               \\
Thickening        & 2526                  & 505               \\
Effusions         & 4608                  & 922               \\
Edema             & 1775                  & 355               \\
Aspiration        & 1543                  & 308               \\
Consolidation     & 1828                  & 366               \\
Mass              & 1403                  & 281               \\
Pneumonia         & 1367                  & 273               \\
Metastatic        & 1283                  & 256               \\
Lymphadenopathies & 1557                  & 311               \\
Hypertension      & 1048                  & 209               \\
Metastases        & 1137                  & 227               \\
Lesions           & 1648                  & 330               \\
Malignancy        & 1070                  & 214               \\
Node              & 850                   & 169               \\
Others            & 3302                  & 661              
\end{tabular}
\caption{Distribution of anomaly counts in the INSPECT dataset for both the training and test sets.}
\label{tab:inspect_count}
\end{table}

\subsubsection*{INSPECT}
The INSPECT dataset has 23,248 scans focusing mainly on pulmonary embolism. After removing some redundant data, we used 17,730 and 3,506 scans for training and testing, respectively. As the dataset is not accompanied by additional abnormality labels and an official split, ReXKG~\cite{zhang2024uncovering} is used to extract entities representing abnormalities. These extracted entities are further utilized for two purposes: (a) stratified train and test dataset split, and (b) multi-anomaly classification to obtain classification logits. After applying ReXKG, the obtained entities are sorted in descending order based on their frequency of occurrence. Entities occurring more than 1,000 times across both training and testing sets are selected for multi-anomaly classification. Also, anomalies with the same Concept Unique Identifier (CUI) are combined, with their frequencies summed accordingly. For example, \textit{emboli, embolus, embolism} have a common CUI \textit{C1704212}, therefore, their occurrence frequencies are aggregated under one entity \textit{``Embolism"}. A list of these anomalies with their occurrence count is shown in Table \ref{tab:inspect_count}. This distribution is used to create a training and test split of the INSPECT dataset via stratified sampling based on the occurrence of anomalies in the reports.

%
%


\begin{algorithm}[H]
\caption{Anomaly-Aware Projection Network Training}
\begin{algorithmic}[1]
\renewcommand{\algorithmicrequire}{\textbf{Input:}}
\Require Image embeddings $\mathbf{z}_i$ obtained using $f_e$, text embeddings $\hat{\mathbf{R}}_i$ and classification logits $\mathbf{l}_i$
\Ensure Frozen LLaMA-3.2-1B with $f_m$ projection network for report generation
\For{each batch $(\mathbf{z}_{i}, \hat{\mathbf{R}}_{i}, \mathbf{l}_i) \in D$}
    \State Concatenate $\mathbf{z}_{i}$ with $\mathbf{l}_{i}$: $\hat{\mathbf{z}}_i \leftarrow \text{concat}(\mathbf{z}_i, \mathbf{l}_i$), where $\hat{\mathbf{z}}_i \in \R^{v+c}$
    \State  Construct prefix mask:  
   $\mathbf{p}_{mask} \leftarrow \mathbf{\mathbf{1}}^{B \times L_p}$, where $B$ is the batch size and $L_p$ is the prefix length (10 in our case)
   \State Concatenate prefix mask with $\hat{\mathbf{R}}_i$ attention mask (obtained using LLaMA-3.2-1B tokenizer):  
   $\text{mask} \leftarrow \text{concat}(\mathbf{p}_{mask}, \text{$\hat{\mathbf{R}}_i$.attention\_mask}, \text{dim}=1)$
    \State Pass $\hat{\mathbf{z}}_{i}$ through the linear projection layer in $f_m$:  $\mathbf{E}_{proj} \leftarrow \mathbf{W}\hat{\mathbf{z}}_i+\mathbf{b}$, where $\mathbf{E}_{proj}\in\mathbb{R}^{L_p\times h}$, $\mathbf{W}\in \mathbb{R}^{(v+c)\times (L_p.h)} $, $\mathbf{b}$ is the bias term and $h$ is the LLM's hidden size.
    \State Reshape($\mathbf{E}_{proj}, B, L_p, h)$, $\mathbf{E}_{proj_{reshape}} \in \mathbb{R}^{B\times L_p \times h}$
    \State Define the learnable prefix constant: $\mathbf{p}_c$, $\mathbf{p}_c \in \mathbb{R}^{B \times L_p \times h}$
    \State Concatenate $\mathbf{\mathbf{E}}_{proj_{reshape}}$ with $\mathbf{p}_c$: $\mathbf{S} \leftarrow \text{concat}(\mathbf{E}_{proj_{reshape}}, \mathbf{p}_c, \text{dim=1})$
    \State Pass $\mathbf{S}$ through $K$ transformer layers (here $K=8$). For each transformer layer, process through multi-head attention and an MLP feedforward network.
    \State Compute loss $\mathcal{L}$ (Eq. \ref{eq:loss})
    \State Backpropagate and update only $f_m$ parameters
\EndFor
\State \Return fine-tuned model for report generation
\end{algorithmic}
\label{algo}
\end{algorithm}


\begin{table*}[ht]
\centering
\resizebox{0.7\linewidth}{!}{
\begin{tabular}{c|c|c|cc}
 Method   & LLM                   & \begin{tabular}[c]{@{}c@{}}No. of \\ Parameters\\ in LLM\end{tabular} & F1-RadGraph & RaTEScore \\ \midrule

                          R2GenGPT  & BioGPT-Large          & 1.6B                                                                                & 0.0208 &	0.3338                                                             \\
                          R2GenGPT & LLaMA-3.2-1B-Instruct & 1B                                                                                  & 0.2310	& 0.6370                                                                     \\
                          R2GenGPT & LLaMA-2-7b-chat-hf          & 7B                                                                                & 0.2783	& 0.6746                                                           \\
                          
                         \textbf{RAD3D-Prefix}     & LLaMA-3.2-1B          & 1B                                                                                  & \textbf{0.2884}    & \textbf{0.6830}                                                              \\ \hline
                                            
\end{tabular}
}
\caption{Comparison with the state-of-the-art approach of vision and text embedding alignment and different sized LLM training based on two more clinically-oriented metrics. }
\label{tab:metrics_appendix}
\end{table*}

\section*{Qualitative Results}
\begin{figure*}[h!]
    \centering
    \includegraphics[width=\linewidth]{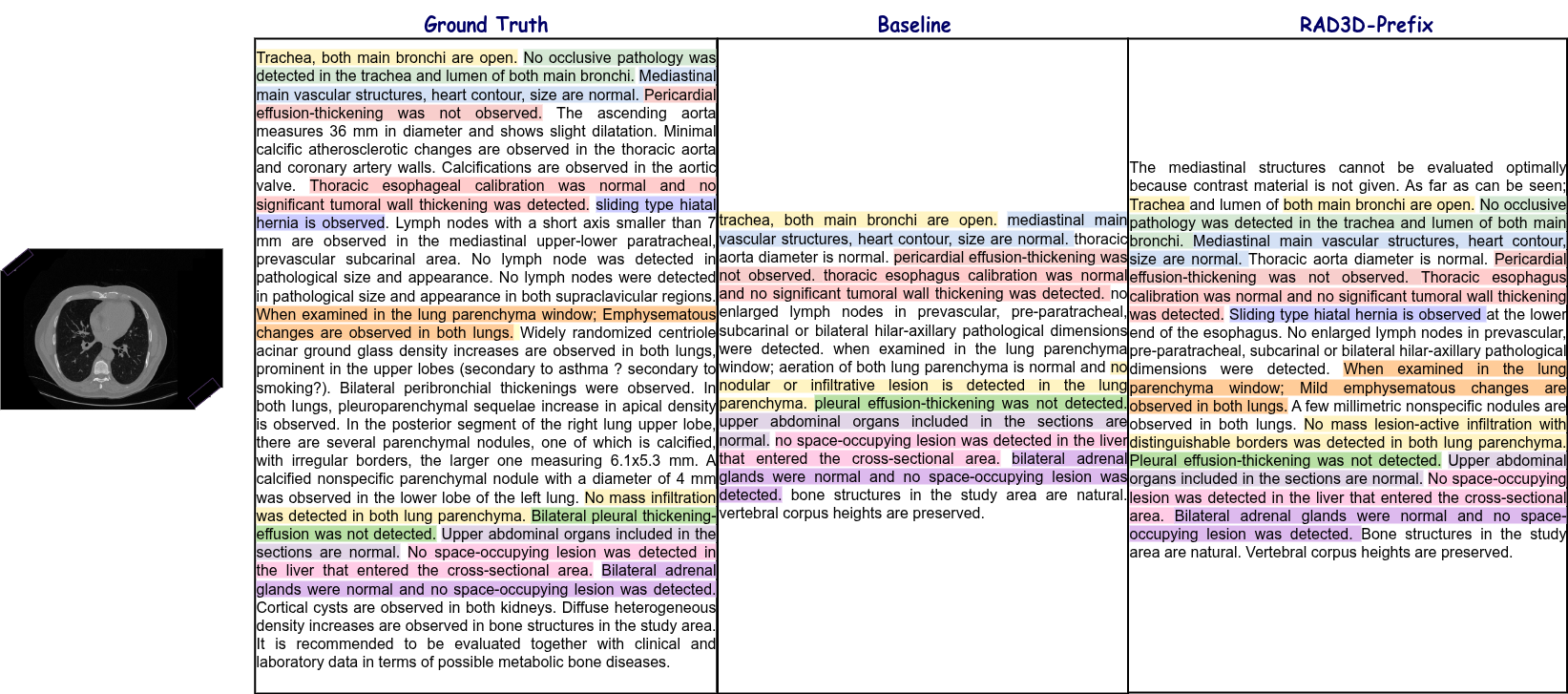}
    \includegraphics[width=\linewidth]{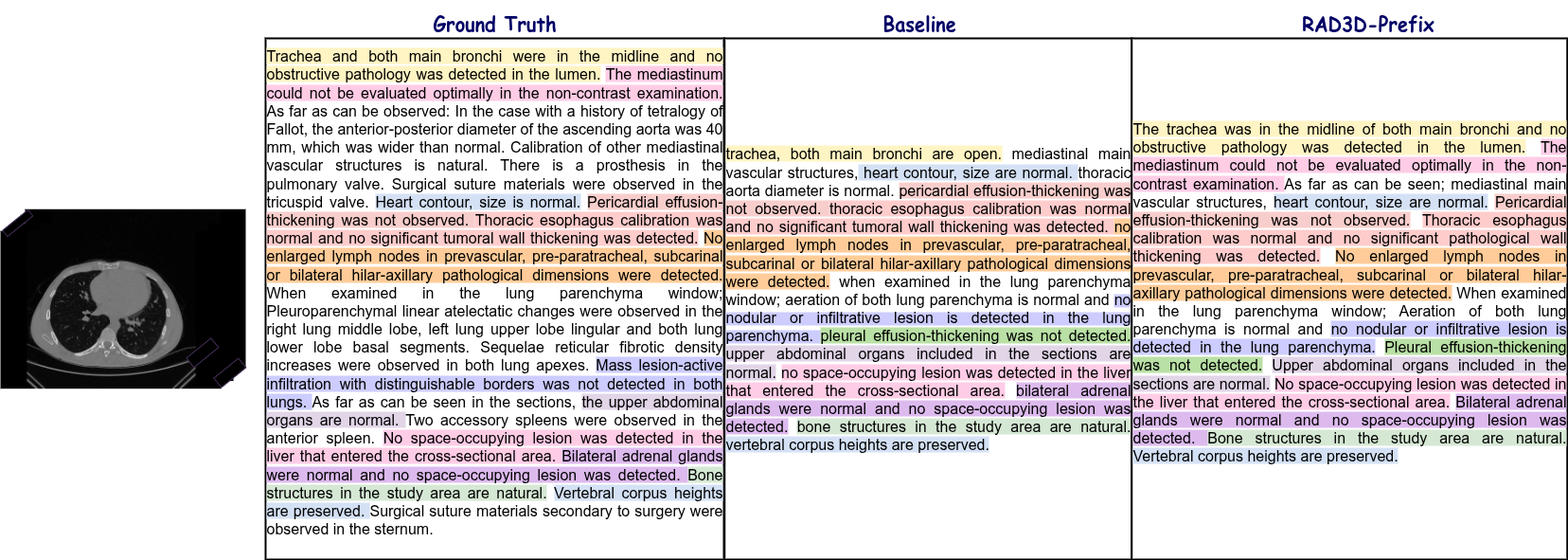}
    \caption{More qualitative examples of the baseline and our proposed method. Matching sentence pairs are highlighted in the same color.}
    \label{fig:app_qual}
\end{figure*}

More qualitative samples are given in Fig. \ref{fig:app_qual}. It can be seen that the baseline model is overfitted on the dataset, generating the same result regardless of the input. In contrast, our model performed better with input-specific output covering most of the anomalies. Some failure cases of our model involved content where abnormalities were accompanied by specific measurements. Nevertheless, our model correctly identified critical conditions, including effusions and hiatal hernia. 

\begin{figure}
    \centering
    \includegraphics[width=\linewidth]{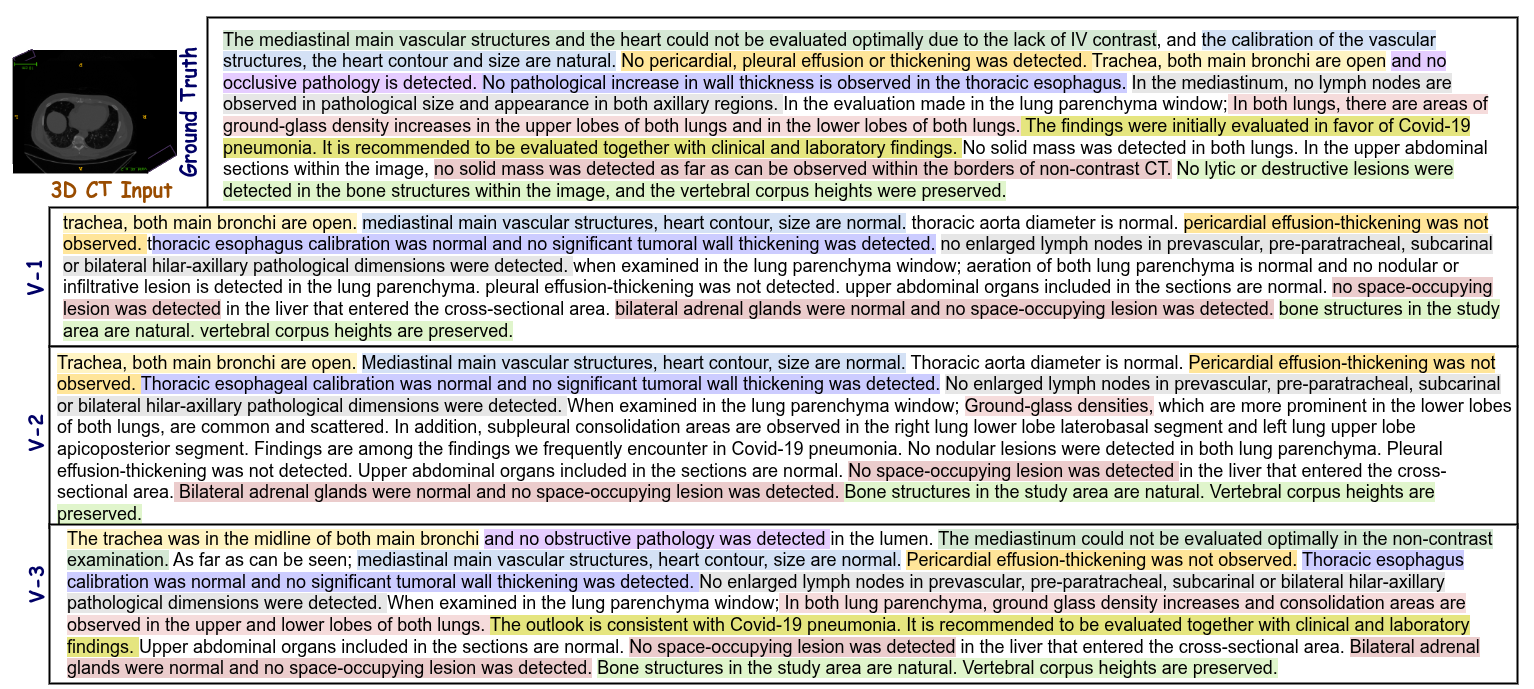}
    \caption{Qualitative sample comparing outcomes of the three variants, \textit{V-1}, \textit{V-2}, and \textit{V-3}.}
    \label{fig:qual_variant}
\end{figure}

Fig. \ref{fig:qual_variant} illustrates reports predicted by the three variants, \textit{V-1}, \textit{V-2}, and \textit{V-3}. It can be observed that RAD3D-Prefix, based on \textit{V-3} variant, roduces the report most aligned with the radiologist-annotated ground truth. It is the only variant that correctly identifies ``COVID-19 pneumonia", ``no obstructive pathology", ``ground-glass opacity with correct location". Following \textit{V-3}, \textit{V-2} performed better than {V-1}, capturing ``ground-glass opacity" but with incorrect location, while \textit{V-1} fails to detect it entirely.  

\section*{GREEN Score Definition and Computation Details}
\begin{figure}[h!]
    \centering
    \includegraphics[width=\linewidth]{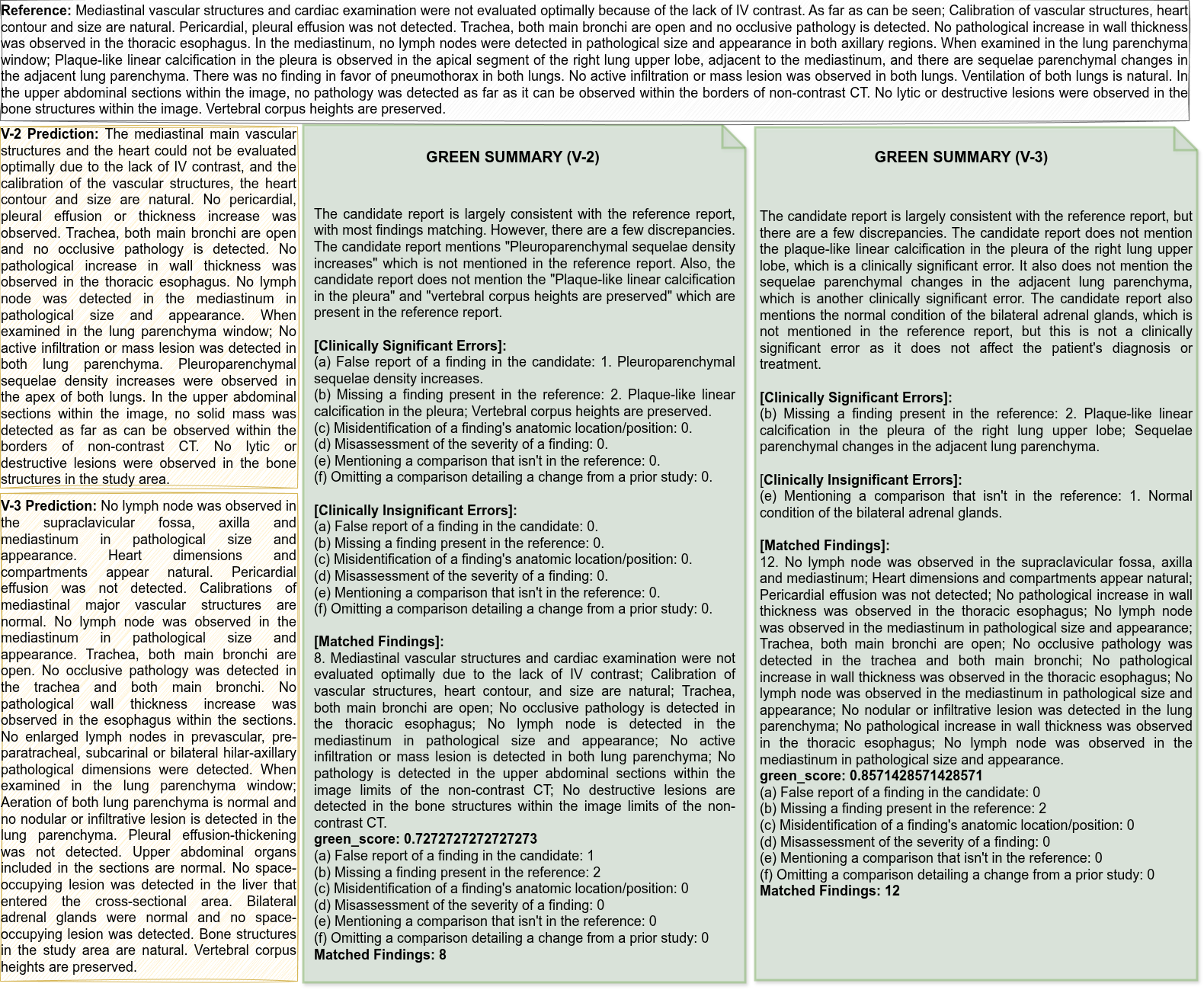}
    \caption{Samples of GREEN Summary for V-2 and V-3 variants.}
    \label{fig:greensummary}
\end{figure}

GREEN Score~\cite{ostmeier-etal-2024-green} is a metric for radiology report generation that uses regular expressions to parse errors in generated reports and to identify matched findings. The score can be calculated as: 
\begin{equation}
\mathrm{GREEN \text{ } Score}  =
\frac{
    \#\,\text{matched findings}
}{
    \#\,\text{matched findings} +
    \sum_{i \in \text{sig.\ errors}} \#\,\text{errors}_{i}
},
\label{eq:green}
\end{equation}

where a ``matched finding'' is a clinical observation present in both the
generated and reference reports. `` errors'' correspond to
findings whose omission or hallucination would plausibly impact clinical
decision-making. If no matched findings are present, the GREEN score is
defined to be zero. Sample GREEN Summaries for \textit{V-2} and \textit{V-3} variants are shown in Fig. \ref{fig:greensummary}. These two variants are compared specifically because \textit{V-3} incorporates classification logits while \textit{V-2} does not, enabling a direct comparison of the impact of logits on clinical correctness. Across 3039 test cases, \textit{V-3} produces 253 additional matched findings ($1.43\%$ improvement) compared to \textit{V-2}, corresponding to approximately one additional finding every 12 cases. Given the sparsity and clinical importance of radiology findings, this improvement is clinically meaningful.

\section*{Projection Network Parameter Count comparison}
\label{sec:parameter}
 To support the effectiveness of our proposed framework, we present results replacing our transformer mapping with a linear layer (see Table \ref{linear}). This variant still uses our proposed prefix design, resulting in increased parameter count from 1.05M (linear) to 1.05M$\times$10 (linear+prefix), still significantly less than 279.5M (original transformer+prefix).
Replacing our transformer mapper (279.5M) with linear layer  cuts parameters 26$\times$ with only a 0.047 BLEU $\downarrow$ (still outperforming R2GenGPT). This shows gains stem from the prefix mechanism, not just increased model size. 10$\times$ extra parameters above account for the prefix length (\textit{our core proposal}).

\begin{figure*}[ht]
    \centering
    \includegraphics[width=0.48\textwidth]{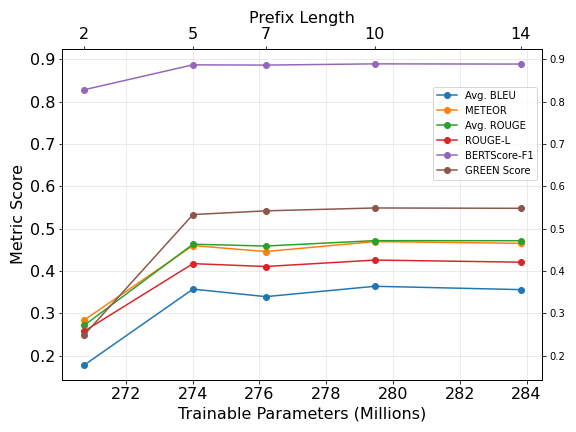}
    \hfill
    \includegraphics[width=0.48\textwidth]{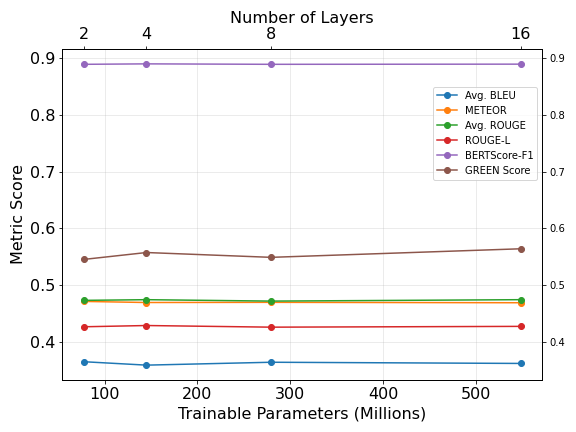}
    \caption{Performance analysis with respect to increasing trainable parameters, influenced by modifications to (a) Prefix Length, (b) Number of Layers in the transformer-based projection network.}
    \label{fig:ablation_prefix}
\end{figure*}

\begin{table}[t]
\resizebox{\columnwidth}{!}{

\begin{tabular}{c|ccccccccc}
\textbf{Method}   & \textbf{\begin{tabular}[c]{@{}l@{}}LLM\\ (1B Frozen \\ Para.)\end{tabular}}                & \textbf{\begin{tabular}[c]{@{}l@{}}Projection\\ (Trainable\\ Para.)\end{tabular}} & \multicolumn{1}{c}{\textbf{\begin{tabular}[c]{@{}c@{}}Avg.\\ BLEU\end{tabular}}} & \textbf{\begin{tabular}[c]{@{}l@{}}METEOR\end{tabular}} & \textbf{\begin{tabular}[c]{@{}l@{}}ROUGE- \\ L\end{tabular}} & \textbf{\begin{tabular}[c]{@{}l@{}}BERTScore- \\ F1\end{tabular}} & \multicolumn{1}{c}{\textbf{\begin{tabular}[c]{@{}c@{}}GREEN\\ (Clinical \\ Efficacy)\end{tabular}}} & \textbf{\begin{tabular}[c]{@{}l@{}}F1- \\ RadGraph\end{tabular}} & \textbf{RaTEScore} \\
\toprule
\textbf{R2GenGPT} & \begin{tabular}[c]{@{}l@{}}LLaMA-3.2-1B\\ -Instruct\end{tabular} & \begin{tabular}[c]{@{}l@{}}Linear Layer\\ (1.05M)\end{tabular}                               & 0.2902                                                                           & 0.3762          & 0.3468                                                      & \textbf{0.8751}                                                      & 0.4120  & 0.2310	& 0.6370                                                                                                    \\
\textbf{Proposed} & \begin{tabular}[c]{@{}l@{}}LLaMA-3.2-1B\end{tabular}             & \begin{tabular}[c]{@{}l@{}}Linear Layer\\ (1.05M $\times$ 10)\end{tabular}                               & \textbf{0.3165}                                                                  & \textbf{0.4161} & \textbf{0.3824}                                             & 0.8714                                                               & \textbf{0.4826} & \textbf{0.2559} &	\textbf{0.6435} \\  
\bottomrule
\end{tabular}
} \caption{Using a linear layer in the projection network instead of the transformer-based network and comparing the setting with R2GenGPT.}
\label{linear}
\end{table}

\section*{Analysis of Prefix Length and Layers Vs. Performance and Trainable Parameters}
We performed an additional ablation study with different prefix lengths (2,5,7, 10, and 14) and the number of layers in the transformer architecture of the projection network  (2, 4, 8, and 16) to analyze tradeoffs between performance and computational overhead. The results are shown in Fig. \ref{fig:ablation_prefix}. It can be observed that increasing prefix length significantly improves performance up to a threshold, after which gains saturate. The most significant improvement occurs between prefix lengths 2 and 5 with gains above 60\% across five metrics. Beyond prefix length 5, the improvements are marginal despite increased parameter count. Further, Fig. \ref{fig:ablation_prefix} (b) shows that while increasing the number of layers dramatically increases trainable parameters, there is no measurable performance gain across NLG metrics. Although a slight improvement is observed in the GREEN metric, the gain is not proportional to the increase in parameters.

\section*{Reader Study}
We conducted a \textit{double-blind reader study} with two clinicians who independently evaluated 100 randomly selected anonymized reports generated by the baseline, \textit{V-2}, and \textit{V-3} models. The evaluation consisted of two sections: clinical accuracy and technical and linguistic quality, each scored on a five-point scale (1–5) according to the following predefined criteria. 
\begin{enumerate}
    \item Clinical Accuracy (1–5)
    \begin{enumerate}
        \item Correct identification of presence/absence of findings
        \item No hallucinated (false positive) findings
        \item Whether errors would meaningfully affect diagnosis or management
        \item Correct anatomical location and placement of findings
    \end{enumerate}
    \item Technical \& Linguistic Quality (1–5)
    \begin{enumerate}
        \item No repetition or redundancy
         \item No incomplete or truncated sentences
         \item No strange symbols or artifacts (e.g., "\#", "?", "Question")

    \end{enumerate}
\end{enumerate}

\textit{Scale: 1 = Very Poor | 2 = Poor | 3 = Acceptable | 4 = Good | 5 = Excellent}

Averaging scores obtained from both clinicians, it is observed that \textit{V-3} improved clinical relevance by 9.8\% over the baseline and by 3.7\% over \textit{V-2}, achieving the highest clinical score. While the baseline achieved higher linguistic fluency scores, our proposed variants produced reports with stronger clinical relevance compared to the baseline. Notably, clinicians observed that the baseline frequently generated normal outcomes, suggesting overfitting and a tendency to prioritize fluent but non-informative text rather than condition-specific findings. Although both variants scored lower than the baseline in linguistic quality, \textit{V-3} still outperformed \textit{V-2 }by 9.4\%.



\bibliography{egbib}
\end{document}